\pdfoutput=1

\documentclass[11pt]{article}

\usepackage
{ACL2023}

\usepackage{times}
\usepackage{latexsym}

\usepackage[T1]{fontenc}

\usepackage[utf8]{inputenc}

\usepackage{microtype}

\usepackage{inconsolata}

\usepackage[capitalize,noabbrev]{cleveref}
\usepackage{hyperref}
\usepackage{booktabs}
\usepackage{graphicx}
\usepackage{xifthen}
\usepackage{xspace}
\usepackage[color=pink!80,textsize=small,textwidth=20mm,colorinlistoftodos]{todonotes}

\newcommand{\tocheck}[2][]{{\color{purple}#2\ifthenelse{\isempty{#1}{}}{}{\todo[color=purple!50]{To check: #1}}}}

\newcommand{\REF}[1][]{\colorbox{red}{\textbf{\$REF\$}}\ifthenelse{\isempty{#1}{}}{\todo[color=red!50]{Missing ref}}{\todo[color=red!50]{Missing ref: #1}}\xspace}

\title{With a Little Help from my (Linguistic) Friends: Topic Segmentation of Multi-party Casual Conversations}

\author{Amandine Decker \\
  Université de Lorraine, CNRS, \\
  Inria, LORIA, F-54000 Nancy, France \\
  University of Gothenburg, Sweden \\
  \texttt{amandine.decker@loria.fr} \\\And
  Maxime Amblard \\
  Université de Lorraine, CNRS, \\
  Inria, LORIA, F-54000 Nancy, France \\
  \texttt{maxime.amblard@univ-lorraine.fr} \\}

\begin{document}
\maketitle

\begin{abstract}
    Topics play an important role in the global organisation of a conversation as what is currently discussed constrains the possible contributions of the participant. Understanding the way topics are organised in interaction would provide insight on the structure of dialogue beyond the sequence of utterances. However, studying this high-level structure is a complex task that we try to approach by first segmenting dialogues into smaller topically coherent sets of utterances. Understanding the interactions between these segments would then enable us to propose a model of topic organisation at a dialogue level. In this paper we work with open-domain conversations and try to reach a comparable level of accuracy as recent machine learning based topic segmentation models but with a formal approach. The features we identify as meaningful for this task help us understand better the topical structure of a conversation.
\end{abstract}

\section{Introduction}

    Topics play a crucial role in understanding and interpreting conversations. When participants have a wrong understanding of the current topic, their contributions can become irrelevant \cite{LogicandConversation} or even incoherent, leading to confusion among the addressees. Similarly, misinterpreting the topic can hinder a participant's ability to understand others' interventions accurately. While topics are more constrained and easily identifiable in controlled settings, such as formal work meetings, open-domain casual conversations have a greater flexibility, allowing participants to switch topics with minimal indication and still be followed by others in the conversation. The larger the number of participants, the more challenging it becomes to maintain control, as everyone contributes to the context.

    Understanding how topics interact in dialogue is thus essential when it comes to modelling dialogue structure beyond the sequence of utterances. However, analysing this structure requires insight on the topics themselves. Being able to segment a dialogue into topically coherent segments seems to be a first step towards modelling high level dialogue structure. The segments could later be linked inside a structure that describes the interactions between them. This task, called dialogue topic segmentation (DTS), finds utility in dialogue generation~\cite{xu-etal-2021-discovering} and summarising~\cite{chen-yang-2020-multi}, among other applications.

    DTS has received less attention compared to monologue or written text topic segmentation, primarily due to the scarcity of annotated data but some DTS approaches get good results on task-oriented dialogues \cite{ijcai2018p612} or conversations with a restricted set of possible topics such as meeting minutes \cite{hsueh-etal-2006-automatic,georgescul-etal-2008-comparative}. \citet{xing-carenini-2021-improving} suggest another method to tackle more varied dialogues. They use the \textit{TextTiling} algorithm \cite{hearst1997texttiling}, that relies on a similarity metric between subsequent blocks of text to identify topic boundaries, and enhance it with a learned utterance-pair coherence scoring model based on BERT \cite{devlin-etal-2019-bert} as similarity metric. They obtain good results in English and Chinese when evaluating their model on three datasets: DialSeg\_711 \cite{Xu_Zhao_Zhang_2021}, Doc2Dial \cite{feng-etal-2020-doc2dial}, and ZYS \cite{Xu_Zhao_Zhang_2021}. Even though these datasets cover different domains, they all contain task-oriented conversations. Evaluating this model on more open-domain dialogues would provide insight on the limits of its generalisation capability. 
    
    In this paper we present an improved version of the original TextTiling algorithm\footnote{Our code is available at \url{https://gitlab.inria.fr/adecker/topicsegmentationtexttiling.git}.}, where we use linguistics properties of dialogue to identify the topic shifts. Our aim is to reach a comparable level of accuracy as the model proposed by \citet{xing-carenini-2021-improving} but with a formal approach. Since we are interested in the structure of topical interactions, an explainable model would help us better understand what features play a role in perceiving topic shifts. A rule-based approach also has the advantage of minimising the amount of computation required by our model, which is more sustainable. Following our goal to build a general model of interaction, we work with multi-party casual conversations, characterised by their more chaotic nature. 

    To summarise our contributions in this work: we  
    (1) reproduced \citet{xing-carenini-2021-improving}'s model;
    (2) trained a Bert-based model to improve the TextTiling algorithm;
    (3) improved the TextTiling algorithm based on linguistic properties;
    (4) evaluated topic segmentation in multiparty casual conversations using the \textit{Friends} corpus.

    


\section{Related work}\label{sec:related_work}

    \subsection{Topic segmentation}

        As explained by \citet{purver2011topic}, while defining a topic may seem straightforward in well-defined tasks such as news broadcasts (each news item), business meetings (agenda items), or court transcripts (arguments), trying to get a finer segmentation can make the task quite complex. Annotators often exhibit disagreement, and finer-grained segmentation leads to even poorer agreement. 

        DTS presents additional challenges compared to monologue topic segmentation. In dialogue settings, interactions create more complex exchanges where the points that are central to the topic under discussion are not necessarily explicit. As a result, producing topic segmentation annotations of great quality is even more complicated and applying technical approaches developed for monologue topic segmentation to DTS is not always successful, these methods are not yet able to tackle open domain conversations \cite{xing-carenini-2021-improving}. 
    
        Existing methods can be broadly categorised into unsupervised techniques (i.e. feature-based approaches) that rely on lexical co-occurrence \cite{hearst1997texttiling,Galley2003segmentation,eisenstein-barzilay-2008-bayesian} or latent topical distribution \cite{eisenstein-barzilay-2008-bayesian,riedl-biemann-2012-topictiling,du-etal-2013-topic} with the assumption that a significant change in vocabulary corresponds to a change in topic \cite{halliday2014cohesion}, and supervised methods\cite{arguello-rose-2006-topic,ijcai2018p612}. However, the lack of annotated dialogue data hinders the progress in neural-based approaches for DTS \cite{hearst1997texttiling}.
    
        One prominent technique used in dialogue topic segmentation is the \textit{TextTiling} algorithm and its extensions. TextTiling was originally introduced by \citet{hearst1997texttiling} and relies on a similarity metric between subsequent blocks of text to identify the topic boundaries. It has been widely employed for topic segmentation in various domains as it is unsupervised. It relies on a similarity metric between subsequent blocks of text to identify the topic boundaries. This method, described in more details in \cref{sec:texttiling}, has been improved in different ways. \citet{galley2003discourse} introduce lexical chains. \citet{Song2016dialogue} use word embeddings to measure the similarity of successive sentences, which is more adapted to dialogue than lexical similarity at a block level. \citet{Xu_Zhao_Zhang_2021,xing-carenini-2021-improving} use BERT \cite{devlin-etal-2019-bert} to capture deeper semantic relations at the utterance level.

        However, these approaches may not be as reliable when applied to casual and open-domain conversations. Multi-party dialogues and long-term conversations add additional complexities to the topic segmentation task. Such conversations can involve multiple simultaneous discussions, references to past conversations that shape the current topic without clear indications, and a shared history among participants that influences the language and references used, potentially deviating from standard usage \cite{yule2013referential}. Additionally, external interruptions by other characters can further disrupt the ongoing topic.
    
        In summary, DTS presents a complex task, due to the inherent chaos introduced by interactions and the scarcity of annotated data. Technical approaches for DTS include feature-based approaches, and neural-based techniques. The adaptations of TextTiling to dialogue and the extensions proposed these past years have shown promising results in the field of dialogue topic segmentation. 
    
        However, further advancements are needed to address the unique challenges posed by open-domain casual conversations and achieve topically coherent segmentation.

    \subsection{TextTiling Approach}\label{sec:texttiling}

\begin{figure}
            \centering
            \includegraphics[width=.48\textwidth]{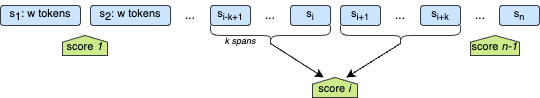}
            \caption{Segmentation in spans of $w$ tokens and computation of the lexical scores in the TextTiling algorithm.}
            \label{fig:TextTiling-Lexical}
        \end{figure}

        \textit{TextTiling} \cite{hearst1997texttiling} is a topic segmentation algorithm that predicts topic boundaries for a given text. It relies on lexical distribution information and its execution follows three main steps: (1) tokenization, (2) lexical score determination, (3) boundary identification. The text is first split in spans of \textit{w} tokens, then a lexical score is computed at the boundary between each span. For instance as represented on \cref{fig:TextTiling-Lexical}, for a text of \textit{n} spans $\{s_1, s_2, ..., s_n\}$, there are $n-1$ boundaries and thus $n-1$ lexical scores to compute.  
    
        Two approaches are suggested in the original paper to measure this score. 
        One is based on the lexical similarity between the two blocks of \textit{k} spans on each side of the boundary. As \cref{fig:TextTiling-Lexical} shows, the lexical score $score(i)$ (corresponding to the boundary $i$) would correspond to the portion of tokens present in both blocks, \textit{i.e}, both in the set of spans $\{s_{i-k+1}, ..., s_{i-1}, s_i\}$ and in the set $\{s_{i+1}, s_{i+2}, ..., s_{i+k}\}$. The other approach focuses on new words in a segment of text. The lexical score $score(i)$ would be the ratio of never-yet-seen words in an interval of $2k$ spans centred around the boundary $i$ divided by the total number of tokens in this interval. Stemming and Lemmatisation are suggested to improve the lexical similarity scores.

\begin{figure}
            \centering
            \includegraphics[width=.48\textwidth]{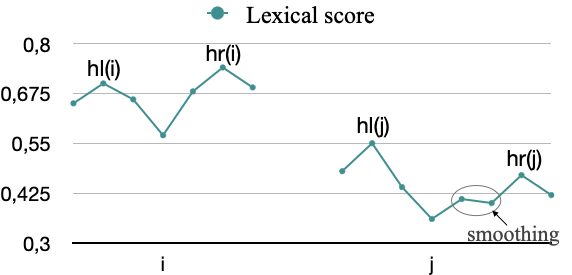}
            \caption{Examples of lexical scores used in the depth scores computation.}
            \label{fig:TextTiling-Depth}
        \end{figure}

        The maximal changes in the lexical scores are then computed thanks to ``depth scores'' by looking at the depth of the ``valley'' in which a given lexical score falls. A deeper valley means that the observed lexical score is more different from previous and later scores, which indicates a higher chance of topic shift. Formally, given a boundary $i$, we measure the depth of this valley by retrieving the first lexical score on the left that forms a pic, \textit{i.e.} $hl(i)$ such that it is greater than the score directly on its left. We retrieve $hr(i)$ similarly on the right. The depth score of the boundary $i$ is then computing by adding the depths on both sides: $dp(i) = \frac{(hl(i) - score(i)) + (hr(i) + score(i))}{2}$. A smoothing of the lexical scores prevents small perturbations to impact the depth computation. \cref{fig:TextTiling-Depth} shows two cases of depth score computation. The first one (i) is classical, where $hl(i)$ and $hr(i)$ are the first pics on the left and right of the considered score. The second case (j) illustrates the role of smoothing, as their is a very small pic between $score(j)$ and the $hr(j)$ we actually consider. Without smoothing, $score(j+1)$ would have been used to compute the depth score while it would not be representative of the real lexical similarity at this point.

        The local maxima of the depth scores are then chosen as topic boundaries. In practice these boundaries are shifted to the closest gap between two paragraphs because the first split into tokens of length $w$ erases this structure. 

\section{Methodology}\label{sec:methodology}


    \subsection{Models}

        We compare two enhancements of the original TextTiling algorithm, one developed by \citet{xing-carenini-2021-improving} based on BERT and one based on linguistic properties. Our goal is to see if a feature-based approach can compete with one based on a language model on complex data such as our \textit{Friends dataset}.

        We use \citet{xing-carenini-2021-improving}'s original dataset for training but also compare the results when adaptating their approach to our dataset. The adaptations and results are detailed in \cref{sec:ML}.

        Our main contribution is the feature-based approach where we adapt the original TextTiling algorithm to dialogue and use more linguistic properties to identify the topic shifts.

    \subsection{Baselines}
    
        As baselines, we use the original TextTiling algorithm that exists in the Python library Nltk\footnote{\url{https://www.nltk.org/_modules/nltk/tokenize/texttiling.html}} as well as the \textit{random} baseline used by \citet{xing-carenini-2021-improving} which assigns boundaries with a probability $\frac{b}{k}$ where $k$ is the number of utterances and $b \in [\![ 0, k-1 ]\!]$ is a randomly chosen number of segment boundaries. We ran ten iterations of this random baseline where only the F$_{k1}$ and F$_{k2}$ had significant differences from one iteration to the other. We thus chose the iteration with the best result on these scores for the final comparisons.

    \subsection{Evaluation Metrics}\label{sec:metrics}

        We use three evaluation metrics for each experiment: $P_k$ error score \citep{beeferman1999statistical}, which is calculated by comparing the model's prediction within a certain sliding window to the ground-truth segments, the standard $F_1$ measure, and a relaxed version of it that we call $F_k$. This adapted F-measure considers that a boundary is correctly identified by the model if there is a ground-truth one at most $k$ utterances before or after the predicted one, the corresponding ground-truth boundary cannot count a second time. In other words, we shift the predicted boundaries that are close to a ground-truth one so that they are considered accurate. We also give twice as much weight to precision compared to recall as we consider that finding right boundaries is more important than finding all of them, which decreases the performance of a model that would suggest boundaries between most utterances.

    \subsection{Dataset}

        For these experiments we use transcriptions in English of the episodes of the TV show \textit{Friends}. Transcripts for all ten seasons (236 episodes), annotated in scenes and with additional notes, were used for the Character Mining project \cite{chen2016character} and their dataset is available online\footnote{\url{https://github.com/emorynlp/character-mining}} (Apache License, Version 2.0).

        Casual conversations are central to human interactions but finding suitable data to analyse them, in particular at the topical level, remains complicated~\cite{gilmartin-campbell-2016-capturing}. For this reason, using transcriptions from a TV show seemed like a good idea for a first approach of our problem as it enabled us to have a sufficient amount of data to work with ML tools, while remaining close enough to real-life dialogues. Studies have indeed shown that spoken language in fiction is quite similar to spontaneous speech \cite{Forchini2009Spontaneity}. 

        The \textit{Friends} dataset is not annotated in topics but we chose to rely on its segmentation in scenes to create the annotations. Additionally, we consider that the notes in the transcripts indicate an important enough change to create a topic shifts. As a result, our assumption is that the topic boundaries coincide with the notes and change in scene. This annotation method is far from perfect but it has the advantage of being objective.

        The example below is an extract of our dataset. We can see that five different characters appear in this short extract, as well as what we consider as two different topics as there is a note between the second and third intervention of this extract. The note explains that a new character enters the room, which is a sufficient disruption to create a topic shift. However in practice, the first speech turn after the note remains on the previous topic and the shift happens right after. This is quite common in the dataset and not unexpected based on real life dialogues, especially when they involve many people. Moreover, the format of the transcriptions is such that concurrent events and/or speech turns are written down in a given order and overlapping are not represented. For this reason, evaluating a topic segmentation solely based on the exact place boundaries should have been placed does not necessarily reflect the quality of a model as we discussed earlier in \cref{sec:metrics}. 
        
        \begin{itemize}
            \setlength\itemsep{-.2em}
            \item[] \textbf{Joey Tribbiani:}	Strip joint! C'mon, you're single! Have some hormones!
            \item[] \textbf{Ross Geller:}	I don't want to be single, okay? I just... I just- I just wanna be married again!
            \item[] \textit{(Rachel enters in a wet wedding dress and starts to search the room.})
            \item[] \textbf{Chandler Bing:}	And I just want a million dollars!
            \item[] \textbf{Monica Geller:}	Rachel?!
            \item[] \textbf{Rachel Green:}	Oh God Monica hi! Thank God! I just went to your building and you weren't there and then this guy with a big hammer said you might be here and you are, you are!
        \end{itemize}

\section{Adapting \citet{xing-carenini-2021-improving}'s BERT-based model to our Dataset} \label{sec:ML}

    \citet{xing-carenini-2021-improving} enhanced the original TextTiling algorithm by replacing the similarity metric by a trained utterance-pair coherence scoring model based on BERT. They use the \textit{Next Sentence Prediction} BERT and fine-tune it with a pairwise ranking loss so that the model learns what pairs of sentences are more or less coherent. They use \textit{DailyDialog} conversations\footnote{\citet{xing-carenini-2021-improving} also trained a model for Chinese on \textit{NaturalConv} but our own work being done with English dialogues we will not discuss this further.} to train their model by feeding it pairs of utterances that they indicate as relatively more or less coherent: Two adjacent utterances (based on Conversation Analysis, \citet{Schegloff1973opening}) are more coherent than two utterances randomly taken from a given conversation (and thus not necessarily adjacent or even subsequent), which in turn are more coherent than two utterances belonging to different conversations. \cref{fig:TextTilingCarenini} in Appendix is a representation of these different levels of coherence. 
        
    This model replaces the original lexical similarity and thus outputs the lexical scores used to compute depth scores and then topic boundaries. It is important to note that the model itself is trained on a pairwise coherence ranking task, which means that it learns to judge how likely two utterances are to follow each other based on the coherence of the pair. The final goal is however to segment a dialogue into topics, the model is thus used with the TextTiling method and evaluated on its ability to produce the valid topic segmentation.

    We applied these enhancements on our own dataset and ran different experiments to assess the performance of the model on multi-party casual conversation such as the ones in the \textit{Friends} dataset.

    We compare the results when \citet{xing-carenini-2021-improving}'s and our data is used for training. We expect better results with our own training data as it would be more similar to the texts we try to segment. 
    
    For our first experiments, we used all seasons except for one as training data and evaluate on the remaining season. However, for a fairest comparison with the feature-based model, which can be evaluated on all seasons, we later worked with models trained on three seasons and evaluated on the seven remaining ones.

        \subsection{Learning Curve}
        Since we evaluate the model on a different task as the one it is trained on, \textit{i.e.,} we evaluate it on the topic segmentation task while it was trained for utterance pair coherence scoring, we wanted to know how much training was needed for the model to show consistent results. We thus trained different models for 10 epochs to see how the results evolved with the training. Even though the loss decreases along training, the results on the actual topic segmentation task do not improve consistently. \cref{tab:ML_results_n3} shows the results for one model and we can see that the evolution in the scores is not consistent but also that the best epoch is not the same for all the measures. Moreover, while for model c-3 (\cref{tab:ML_results_n3}) the best results can be found among the last epochs, other models found in Appendix give better results in their first epochs (\cref{tab:ML_results_s3,tab:ML_results_n1}).

        \begin{table}[ht]
            \centering
            \begin{tabular}{lcccc}
                \toprule
                Experience & F1 $\uparrow$ & F$_{k1}$ $\uparrow$ & F$_{k2}$ $\uparrow$ & P$_k$ $\downarrow$ \\
                \midrule
                Epoch 1 & 19.25 & 44.05 & 51.36 & 50.26 \\
                Epoch 2 & 19.39 & 45.98 & 52.53 & 51.55 \\
                Epoch 3 & 19.53 & 46.14 & 53.49 & 50.44 \\
                Epoch 4 & 17.83 & 41.04 & 49.53 & \textbf{48.97} \\
                Epoch 5 & 20.18 & 46.10 & 53.58 & 51.47 \\
                Epoch 6 & 20.47 & 46.10 & 52.60 & 50.84 \\
                Epoch 7 & 19.66 & 43.91 & 51.64 & 51.94 \\
                Epoch 8 & 20.50 & \textbf{46.76} & \textbf{54.09} & 51.01 \\
                Epoch 9 & 20.53 & 45.34 & 51.87 & 51.51 \\
                Epoch 10 & \textbf{20.73} & 45.90 & 52.93 & 51.71 \\
                \bottomrule
            \end{tabular}
            \caption{Average results of 10 epochs for the model c-3.} 
            \label{tab:ML_results_n3}
        \end{table}

        \subsection{Coherence Layers}
            Regarding the coherence layers, our dataset is not annotated in dialogue acts nor topics, which makes it impossible to use adjacency pairs or utterances from different topics as \citet{xing-carenini-2021-improving} suggest. However, we believe that the annotation in scenes, episodes and seasons, as well as the additional notes in the transcripts, provide sufficient information to build utterance pairs of different coherence. As mentioned earlier, we consider that a note usually indicates an important enough change to create a topic shift, for this reason they are another type of boundary we consider when building our pairs and we subdivide each scene in smaller spans of utterances based on the note boundaries.
            As a consequence, we consider the following types of boundaries in decreasing order of coherence: note, scene, episode, season. In practice, it means that we have six levels of coherence ranked in decreasing order:
            \begin{itemize}
                \setlength\itemsep{-.5em}
                \item[a] subsequent utterances within the same note span;
                \item[n] randomly picked utterances within the same note span;
                \item[c] randomly picked utterances within the same scene;
                \item[e] randomly picked utterances within the same episode;
                \item[s] randomly picked utterances within the same season;
                \item[d] randomly picked utterances within different seasons.
            \end{itemize}
            As \citet{xing-carenini-2021-improving} only had three levels of coherence, we try several settings with our own data. We include the subsequent utterances [a] and the randomly picked utterances within the same note span [n] in all our settings to reproduce the `adjacent' and `same dialogue' coherence levels from \citet{xing-carenini-2021-improving}. We try different layers for the third coherence level (within the same scene [c], within the same episode [e] and within different seasons [s]), and we also train one model with more layers: [c], [e], [s], and [d].
            \cref{tab:ML_results_e0} shows the results of different models after one epoch of training. The models named X-1 were trained on a dataset containing utterances from all the seasons except the first one (and thus evaluated on season 1), while the models named X-3 were trained on all the seasons except the third one. We can see that in both cases, the model [d] shows the best results. However, if we have a look at the results for the second epoch, model [c]-1 gets better results. 
        
            We see again that the results are not consistent throughout the epochs and choosing the best setting in terms of layers of comparisons is complicated.

            However, we can see that having more layers does not seem to provide better results so we decided to work with three layers like \citet{xing-carenini-2021-improving}. We worked with subsequent utterances within the same note span [a], randomly picked utterances within the same note span [n] and randomly picked utterances within the same episode [e]. [a] and [n] to reproduce the first two layers of \citet{xing-carenini-2021-improving} as said before, and [e] because it is the middle coherence layer that we have.

            Another problem of the models we have discussed so far was that they were trained on nine out of the ten seasons of the dataset, which leaves only one season for the evaluation while the feature-based model can be evaluated on all of them. We thus trained some models for two epochs on one, three and four seasons and saw that using only one season produced significantly worse results. We eventually decided to work with three seasons for our comparisons with the feature-based model.

            \begin{table}[ht]
                \centering
                \begin{tabular}{lcccc}
                    \toprule
                    Experience & F1 $\uparrow$ & F$_{k1}$ $\uparrow$ & F$_{k2}$ $\uparrow$ & P$_k$ $\downarrow$ \\
                    \midrule
                    ML [c]-1 & 25.96 & 57.49 & 62.74 & 51.97 \\
                    ML [d]-1 & \textbf{27.18} & \textbf{58.69} & \textbf{63.61} & \textbf{50.20} \\
                    ML [cd]-1 & 21.94 & 51.06 & 62.33 & 53.37 \\
                    ML [cesd]-1 & 25.95 & 56.34 & 61.89 & 50.54 \\
                    \midrule
                    ML [c]-3 & 19.25 & 44.05 & 51.36 & 50.26 \\
                    ML [d]-3 & \textbf{20.42} & \textbf{48.22} & \textbf{54.15} & 54.22 \\
                    ML [cd]-3 & 20.37 & 45.06 & 53.03 & \textbf{47.32} \\
                    \bottomrule
                \end{tabular}
                \caption{Average results for different models trained on \textit{Friends} with different coherence layers (Epoch 1).}
                \label{tab:ML_results_e0}
            \end{table}

        \subsection{Model Stability}
        To assess the stability of our model we trained different versions of it on different training subsets based on the same coherence layers. We built three subsets (based on seasons 2, 3 and 4, and with the coherence layers a, n and e as stated above) and trained three models per subset for two epochs. The results can be seen in Appendix in \cref{tab:instability}. When compared with a t-test, about half of the models gave significantly different results whether the comparison was done between epochs, between models trained on the same training subset or on different subsets. Some models performed well in terms of F-measure (the classical one as well as our adapted version) but worse than the others in terms of P$_k$.


        To do the fairest comparison with our feature-based model, we chose to work with the best version of this model. Since none of the models was performing the best on all of the measures we considered, we chose the best one in terms of P$_k$ among the ones with best F-measures (\textit{d2 m2 e2} in \cref{sec:comparison_all}). While the results seem lower than those of other models we presented in this section (see \cref{tab:ML_results_e0} for example), this model was trained on three seasons (instead of nine for the previous models) and can thus be evaluated on seven seasons (instead of one), which can explain the lower results.

\section{Improving the original TextTiling algorithm with Linguistic Features}\label{sec:FB}
    In parallel to the experiments with the BERT-based model, we worked on enhancing the original TextTiling algorithm with more linguistic features. Such a model has the advantage of being explainable, as opposed the BERT-based one.

    Two basic ideas are discussed in the literature when it comes to identifying changes in topics \citep{purver2011topic}. The first one is that a change in topic implies an important change in terms of content. For example, it corresponds to the introduction of a new vocabulary \cite{youmans1991tool} which is more or less constant inside a topic \cite{morris-hirst-1991-lexical}. Additionally in a dialogue, the most active participants can change based on the topic. The second insight is that there exist distinctive topic boundary features such as discourse markers or aspects of the prosody. Questions can also indicate a continuity of the current topic.

     We decided to include both approaches in our version of TextTiling. Our idea was to complexify the similarity metric by taking more features of the text into account. 

    \subsection{Adaptations of the Original TextTiling Algorithm}

        The original TextTiling algorithm proposes two approaches to segment a text. In the Block Comparison approach, the lexical scores represent the similarity between two blocks in terms of tokens. Two blocks with numerous tokens in common will have a higher score than a block that has unique tokens compared to the other block. The Vocabulary Introduction method focuses on the amount of new tokens in two consecutive blocks compared to the number of non stop-word tokens in the blocks. 
        Instead of considering only never-yet-seen words we use a memory parameter $m$: a word is considered new if it did not appear in the $m$ last sentences of the text. We set this parameter to 20. This adaptation accounts for the fact that in a long dialogue, a topic can be resumed after talking about something else.

        Moreover, the original TextTiling algorithm splits the text in spans of $w$ tokens which we supposed was not the most relevant for dialogue. For this reason, our adaptation splits the text in sequences of utterances such that the number of tokens is the closest possible to $w$ (with $w=12$ as it seems to be a reasonable length for an informative utterance). In practice, most of the spans contain only one utterance, they can contain more when the utterances are very short and are thus less likely to be informative.

    \subsection{New Feature-Based Additions} \label{sec:feature_based}

        We also wanted to consider other features of dialogue when computing the similarity scores. We considered the changes in speakers throughout the conversation \cite{nguyen-etal-2012-sits}. We tried two different ways to modify the depth scores obtained after the block comparison or vocabulary introduction. In one case we increased the depth scores following each utterance that introduced a new speaker. The other modification we tried was inspired from the Block Comparison method. We computed a depth score for each speaker of the conversation based on their proportion of interventions in a block. It means that on top of considering great changes in the lexicon (original Block Comparison method), we also consider changes in terms of speaker distribution (speaker depth scores). The mean of all these scores was then averaged with the original depth scores, where the original scores weight twice as much as the speaker scores. 

        We took questions into considerations with the assumption that a topic shift would not directly follow a question. This hypothesis is rather naive but we decided to see what results a very basic implementation could produce.

        And lastly we used the \textit{coreferee} Python library to take coreference chains \cite{schnedecker1997nom} into account in the computation of our depth scores. Our assumption was that a topic shift is less likely to exist inside a coreference chain. For this reason, we smoothed the gap scores between the first and last mention of a given reference.

        \subsection{Comparison of the Features}

        We tried these features separately and combined them in different ways to see which combinations would give us the best results. The experiments are summarised in \cref{tab:feature_based_results}.
        
        \begin{table}[ht]
            \centering
            \begin{tabular}{p{5.8em}cccc}
                \toprule
                Experience & F1 $\uparrow$ & F$_{k1}$ $\uparrow$ & F$_{k2}$ $\uparrow$ & P$_k$ $\downarrow$ \\
                \midrule
                BC & 10.78 & 30.57 & 45.60 & 49.58 \\
                VI & 9.78 & 27.55 & 43.29 & 52.40 \\
                BC+VI & 11.00 & 30.50 & 45.85 & 49.51 \\
                BC+SI & 10.89 & 29.95 & 46.14 & 48.57 \\
                BC+SD & \textbf{14.94} & \textbf{39.55} & \textbf{52.38} & \textbf{46.70} \\
                BC+SD+Q & \textbf{14.94} & \textbf{39.55} & \textbf{52.38} & \textbf{46.70} \\
                BC+SD+S & \textbf{14.94} & \textbf{39.55} & \textbf{52.38} & \textbf{46.70} \\
                BC+VI+Co+SD & \textbf{15.45} & \textbf{40.32} & \textbf{53.63} & \textbf{47.43} \\
                \bottomrule
            \end{tabular}
            \caption{Results of different feature based models. \\ \textit{Block Comparison (BC), vocabulary introduction (VI), coreference chains (Co), speaker introduction (SI), speaker depth (SD), questions (Q), stemming (S)}}
            \label{tab:feature_based_results}
        \end{table}

        The best results are obtained with the Block Comparison method augmented by the Speaker Depth feature. The results are equivalent to the model that combined Block Comparison, Vocabulary Introduction, Coreference chains and Speaker Depth. However, coreference chains are computationally expensive to retrieve, which makes the model BC + SD more interesting. 
        
        We can also see that stemming the text does not improve the results. Lemmatisation gave the same result. This could be due to the data being artificial in the sense that scenarists may avoid repetitions when it is not for the sake of humour.
        
        In the following, we will hence use BC + SD for comparisons with other models.

\section{Final Comparisons and Conclusion}\label{sec:ccl}

    \subsection{Comparison of All the Models}\label{sec:comparison_all}

        \cref{tab:all_results} summarises the results for the best feature-based model, the best ML-based model trained on \textit{Friends} and the best ML-based model trained on the original data from \citet{xing-carenini-2021-improving}, as well as the two baseline models discussed above (random baseline and original TextTiling algorithm). These results are based on the evaluation on season one and five to ten only as the ML-based model was trained on the seasons two to four.
        
        \begin{table}[ht]
            \centering
            \begin{tabular}{lcccc}
                \toprule
                Experience & F1 $\uparrow$ & Fk1 $\uparrow$ & Fk2 $\uparrow$ & P$_k$ $\downarrow$ \\
                \midrule
                BC + SD & 15.06 & 40.36 & \textbf{52.94} & \textbf{46.44} \\                    
                ML \textit{Friends} & \textbf{18.98} & \textbf{42.71} & 48.45 & 48.43 \\
                ML OG data & 15.07 & 38.33 & 47.79 & 56.41 \\
                OG TextTiling & 10.90 & 32.43 & 46.90 & 52.45 \\
                Random & 13.95 & 37.74 & 42.28 & 55.74 \\
                \bottomrule
            \end{tabular}
            \caption{Comparison of the different models (Evaluation: Seasons 1, 5, 6, 7, 8, 9, 10 only).}
            \label{tab:all_results}
        \end{table}

        As we expected, using the \textit{Friends} dataset for training gives significantly better results than a less relevant dataset, as the one used originally by \citet{xing-carenini-2021-improving}. Nevertheless, we can note that in terms of F1-score, our feature-based model and the ML-based model trained on the original data are equivalent. As we have explained earlier, the F1-score is not the most meaningful measurement for the topic segmentation task but this result still shows a certain generalisation capacity from \citet{xing-carenini-2021-improving}'s model.

        We also see a clear improvement between the original TextTiling algorithm and our enhanced version, especially for the P$_k$, which shows that the linguistic properties we considered and described in \cref{sec:feature_based} are relevant for our task.

        The best model is the ML-based model trained on \textit{Friends} when we look at the F1 and the F$_{k1}$. However, our feature-based model is better in terms of P$_k$ and F$_{k2}$. This shows that for the Topic Segmentation task a feature-based model can compete with language models on certain types of dialogues. Moreover, the BERT-based model is not very stable on our dataset, which we believe is due to the complexity of multi-party casual conversations as opposed to the more controlled dialogues usually used in Topic Segmentation. Our approach based on linguistic features provides an explainable baseline.

        \subsection{Conclusion and Future Work}

    In this paper we investigated the task of linear topic segmentation on multi-party casual conversations. Since this kind of data is complicated to obtain, we chose to work on transcriptions of the TV-show \textit{Friends} as this dataset is available online. The number of speakers and the context of the dialogues creates the possibility for various types of topic shifts which can be challenging for a model. We used the TextTiling approach which uses a similarity metric between subsequent parts of a text to identify the topic shifts. We enhanced it with more linguistic properties that could play a role in identifying topic shifts, and compared it to the same approach but enhanced with a trained utterance-pair coherence scoring model based on BERT.


    As BERT has been trained on the next sentence prediction task, it seems like a relevant model for topic segmentation and in particular to improve the TextTiling approach. Other similar models such as BART \cite{lewis-etal-2020-bart} or T5 \cite{Colin2020T5} did not seem as suitable for our work as they have not been explicitly trained for the next-sentence prediction task. The generalisation capabilities of T5 would probably make it able to produce similar results to BERT, or even better ones, but it would be more complicated to understand the dialogue features used to identify topic shifts. These reasons explain why we chose to use BERT, as \citet{xing-carenini-2021-improving} had done.
    
    While the BERT-based improved model showed good results, it did not significantly outperform the enhanced feature-based approach with all the measures we considered. It would be interesting in a future work to see if T5 or newer models produce better results. Our concern on explainability was however central in this first set of experiments. For this reason, working on improving even more our feature-based approach by investigating the different types of topic shifts and their linguistic specificities could be very insightful. It would provide us more clues on the structure of interaction and help us create a model of it at the topic level.

\section*{Limitations}


In this study, we used the model BERT for one aspect of our experiments. We acknowledge that this model is not the most recent one but we considered it suitable for our task thanks to its specific training for next-sentence prediction. Working with more recent models would imply a higher energy cost while we believe these models would lack the explainability we are looking for in terms of structure.

We also chose to work on transcriptions from fictional dialogues, which creates two limitations. We discussed one of them in the paper when we explained that the fictional aspect of these conversations was likely not the source of huge differences with natural casual conversations. The second limitation however concerns the lack of multi-modality of our work. Transcriptions cannot contain all the information (visual, prosodic, etc.) required to capture fully a conversation. In particular, our dataset did not contain any prosodic information and lacked most of the visual context one may need to understand topic shifts that rely on a change in the context. While the notes could have brought some additional information, we chose to focus on linguistic information in this study. But future work on topic identification should include more modalities to be complete.

\section*{Ethical Statement}

For this experiment, we did not employ any people and we used tools that were free to use.

We have taken care to ensure that the data used is representative of a certain diversity. For example, the corpus is the corpus is balanced in terms of gender. However, we acknowledge that working with the TV show \textit{Friends} covers little cultural diversity.

\bibliographystyle{acl_natbib}
\bibliography{anthology,custom}

\newpage
\appendix

\section{Appendix}
\label{sec:appendix}

    \subsection{Computational Resources Used}

        We tried to limit our use of heavy computational powers. Our feature-based model was run on a local machine and except for the experiments that involved co-reference chains identification, creating the topic segmentation of one episode of \textit{Friends} takes less than a few seconds.

        As for the experiments using Machine Learning, we did our best to optimise the batch sizes and the number of experiments we could run in parallel to reduce the training time as much as possible. We ran our experiments on the Lark servers from CLASP (Gothenburg University) where we used one Nvidia Titan RTX GPU. Our model is based on the \textit{Next Sentence Prediction} BERT model \cite{devlin-etal-2019-bert}, each epoch took about one hour of training.

    \subsection{Different Coherence Levels Considered by \citet{xing-carenini-2021-improving}}

\begin{figure}[h!]
            \centering
            \includegraphics[width=.4\textwidth]{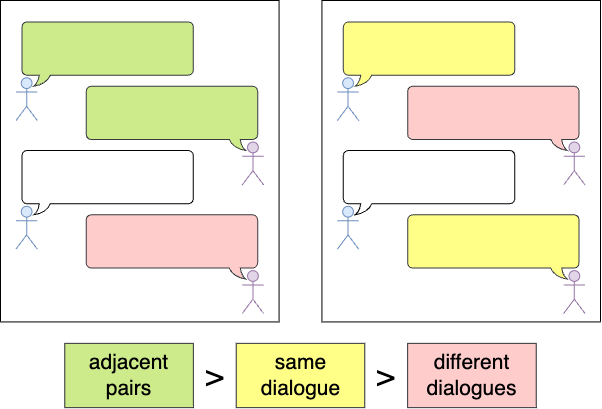}
            \caption{Levels of coherence considered by \citet{xing-carenini-2021-improving}}
            \label{fig:TextTilingCarenini}
        \end{figure}

        \cref{fig:TextTilingCarenini} illustrates the three levels of coherence \citet{xing-carenini-2021-improving} used in their experiment. As explained above, we had the possibility to use more different layers thanks to the segmentation in notes of our dataset.
    
    \newpage
    \subsection{Additional results of the Machine-learning-based Approach}

        \paragraph{Learning Curve} \cref{tab:ML_results_s3,tab:ML_results_n1} show the results when training one model on nine out of the ten available seasons of \textit{Friends} for ten epochs. We can see that the results do not consistently improve from one epoch to another and the differences between each epochs are not very big. A t-test indicates that the results are significantly different from one epoch to another, however the evaluation set is small due to the fact that these models were trained on nine seasons. Hence, we decided to stop training our models for such a long time and trained them for only two epochs in our later experiments.

        \begin{table}[h]
            \centering
            \begin{tabular}{lcccc}
                \toprule
                Experience & F1 $\uparrow$ & F$_{k1}$ $\uparrow$ & F$_{k2}$ $\uparrow$ & P$_k$ $\downarrow$ \\
                \midrule
                Epoch 1 & 20.42 & \textbf{48.22} & \textbf{54.16} & 54.22 \\
                Epoch 2 & 19.92 & 44.50 & 51.38 & 53.09 \\
                Epoch 3 & 18.90 & 43.76 & 51.68 & \textbf{51.40} \\
                Epoch 4 & 19.91 & 46.13 & 52.11 & 52.54 \\
                Epoch 5 & 19.96 & 47.19 & 53.59 & 53.49 \\
                Epoch 6 & 19.81 & 46.08 & 51.69 & 52.67 \\
                Epoch 7 & 20.54 & 46.49 & 53.09 & 53.71 \\
                Epoch 8 & \textbf{20.75} & 47.23 & 52.47 & 53.39 \\
                Epoch 9 & 20.50 & 46.88 & 52.44 & 53.93 \\
                Epoch 10 & 20.14 & 47.00 & 52.44 & 54.18 \\
                \bottomrule
            \end{tabular}
            \caption{Resutls of 10 epochs for the model d-3.} 
            \label{tab:ML_results_s3}
        \end{table}
    
        \begin{table}[h]
            \centering
            \begin{tabular}{lcccc}
                \toprule
                Experience & F1 $\uparrow$ & Fk1 $\uparrow$ & Fk2 $\uparrow$ & P$_k$ $\downarrow$ \\
                \midrule
                Epoch 1 & 25.96 & 57.49 & 62.74 & 51.97 \\
                Epoch 2 & \textbf{27.48} & \textbf{59.19} & \textbf{63.84} & 50.71 \\
                Epoch 3 & 26.31 & 58.50 & 62.81 & 52.44 \\
                Epoch 4 & 27.18 & 58.03 & 63.03 & 51.90 \\
                Epoch 5 & 27.42 & 58.01 & 62.71 & 51.37 \\
                Epoch 6 & 27.09 & 58.50 & \textbf{63.89} & 51.12 \\
                Epoch 7 & 26.53 & 57.24 & 62.44 & \textbf{50.56} \\
                Epoch 8 & 26.51 & 57.72 & 62.50 & 52.04 \\
                Epoch 9 & 26.72 & 57.71 & 62.83 & 51.51 \\
                Epoch 10 & 26.81 & 57.54 & 62.96 & 51.61 \\
                \bottomrule
            \end{tabular}
            \caption{Results of 10 epochs for the model c-1.}
            \label{tab:ML_results_n1}
        \end{table}

        \newpage
        \paragraph{Model Stability}

        To assess the stability of our model we trained different versions of it on different training subsets based on the same coherence layers. We built three subsets (seasons 2, 3 and 4, coherence layers a, n and e) and trained three models per subset for two epochs. \cref{tab:instability} shows that about half of the models gave significantly different results when compared with a t-test whether the comparison was done between epochs, between models trained on the same training subset or on different subsets. Some models performed well in terms of F-measure but worse than the others in terms of P$_k$.
    
        \begin{table}[ht]
            \centering
            \begin{tabular}{lcccc}
                \toprule
                Experience & F1 $\uparrow$ & F$_{k1}$ $\uparrow$ & F$_{k2}$ $\uparrow$ & P$_k$ $\downarrow$ \\
                \midrule
                d1 m1 e1 & \textbf{17.66} & \textbf{44.82} & \textbf{48.66} & 54.22 \\
                d1 m1 e2 & \textbf{17.91} & \textbf{44.97} & \textbf{49.04} & 54.00 \\
                d1 m2 e1 & 12.26 & 31.69 & 42.09 & 56.93 \\
                d1 m2 e2 & 12.76 & 33.48 & 43.20 & 56.95 \\
                d1 m3 e1 & 15.35 & 38.27 & \textbf{47.30} & 53.10 \\
                d1 m3 e2 & 14.78 & 38.16 & \textbf{47.52} & 53.35 \\
                \midrule
                d2 m1 e1 & 16.92 & 39.59 & 45.94 & 50.78 \\
                d2 m1 e2 & 16.20 & 35.42 & 42.78 & 50.21 \\
                d2 m2 e1 & \textbf{18.94} & \textbf{42.93} & \textbf{48.57} & 49.21 \\
                d2 m2 e2 & \textbf{18.98} & \textbf{42.73} & \textbf{48.47} & 48.43 \\
                d2 m3 e1 & \textbf{18.36} & 41.99 & \textbf{47.58} & 49.55 \\
                d2 m3 e2 & \textbf{18.67} & 41.37 & \textbf{47.44} & 48.54 \\
                \midrule
                d3 m1 e1 & \textbf{17.43} & 36.53 & 43.79 & \textbf{46.88} \\
                d3 m1 e2 & \textbf{18.74} & 38.15 & 44.52 & \textbf{45.63} \\
                d3 m2 e1 & 14.61 & 34.03 & 42.33 & 50.58 \\
                d3 m2 e2 & 10.70 & 28.29 & 42.97 & 54.82 \\
                d3 m3 e1 & \textbf{17.94} & \textbf{43.75} & \textbf{48.87} & 52.44 \\
                d3 m3 e2 & \textbf{18.10} & \textbf{42.70} & \textbf{48.11} & 51.23 \\
                \bottomrule
            \end{tabular}
            \caption{Results of different models trained on three training subsets (d1, d2, d3) for two epochs each.\vspace{-1em}}
            \label{tab:instability}
        \end{table}

\end{document}